\documentclass[letterpaper, 10 pt, conference]{ieeeconf}  

\IEEEoverridecommandlockouts                              

\usepackage{amsmath} 
\usepackage{amssymb}  
\usepackage{graphicx}
\usepackage{booktabs}
\usepackage{multirow}
\usepackage{tikz}
\usepackage{svg}
\usepackage{float}
\usepackage{xcolor}
\usepackage{wrapfig}
\usepackage[absolute,overlay]{textpos}
\usetikzlibrary{shapes.geometric, arrows.meta, positioning, fit, backgrounds, calc}

\makeatletter
\let\NAT@parse\undefined
\makeatother
\usepackage{hyperref}

\title{\LARGE \bf
Evolutionary Discovery of Developmental Reward Schedules in Deep Reinforcement Learning
}

\author{Alan Nadelsticher Ruvalcaba$^{1,2}$ 
\thanks{\copyright~2026 IEEE. Personal use of this material is permitted.  Permission from IEEE must be obtained for all other uses, in any current or future media, including reprinting/republishing this material for advertising or promotional purposes, creating new collective works, for resale or redistribution to servers or lists, or reuse of any copyrighted component of this work in other works.}%
\thanks{$^{1}$Alan Nadelsticher Ruvalcaba is with the College of Computing at the Georgia Institute of Technology, Georgia, USA; and $^{2}$with the School of Engineering and Applied Science at the University of Pennsylvania, Pennsylvania, USA. Email:
        {\tt\small alan.nadelsticher@gatech.edu}}%
}

\begin{document}

\maketitle
\begin{textblock*}{\textwidth}(0.75in,10.50in)
\centering
\footnotesize\itshape Accepted at the 2026 IEEE International Conference on Development and Learning (ICDL)
\end{textblock*}

\thispagestyle{empty}
\pagestyle{empty}

\begin{abstract}
The temporal structure of reward composition in reinforcement learning (RL) is typically hand-designed and held fixed throughout training, leaving the progression of motivational priorities largely unexplored. In this work, we propose an evolutionary framework for discovering developmental reward schedules, in which three distinct biologically inspired motivational components --- agency, novelty, and reactivity --- are combined through time-varying weights that dynamically shift over the course of training. Evaluated on two sparse-reward MiniGrid tasks: DoorKey-6x6 and KeyCorridorS3R1, our framework compares the generalizability of four evolutionary algorithms: CMA-ES, xNES, DE, and L-SHADE against an extrinsically motivated baseline (our main comparison point), and three additional hand-designed methods. On DoorKey-6x6, all evolved methods outperform the non-evolved baselines, with L-SHADE achieving the best performance --- an approximate relative mean improvement of 11.4\% over the extrinsic only baseline. On KeyCorridorS3R1, CMA-ES achieves the best overall performance, with the remaining evolved methods showing weaker and less reliable generalization capability compared to the extrinsic only baseline. Interestingly, the discovered schedules diverge from our defined developmental ordering, with novelty consistently emerging as the dominant early signal during training, across both tasks. Collectively, our results position evolutionary optimization as a promising approach for developmental reward schedule discovery in deep reinforcement learning, and suggest that what evolution finds to be optimal in computational settings may differ from what it finds to be optimal in biology. The code for this project can be found at: \url{https://github.com/alannadels/Evolutionary_RL.git}.
\end{abstract}

\section{Introduction}

In humans, and other mammals alike \cite{luciana2012}, the dopaminergic reward system is driven by a shifting ensemble of non-arbitrary, extrinsic and intrinsic motivational signals throughout development: neonates first discover behavior-event contingency \cite{heckhausen2010}, infants then develop novelty preferences for stimuli of intermediate complexity \cite{kidd2012}, adolescents exhibit a hyper-responsive striatal reward system that peaks before prefrontal control matures \cite{galvan2010, cohen2010, casey2008} and, finally, mature goal-directed behavior emerges only once corticostriatal connectivity develops in early adulthood \cite{insel2017}. This progression reflects a functionally optimal strategy for motivational development --- one that evolution itself discovered over time.

Yet, most reinforcement learning (RL) systems treat the temporal structure of reward composition as a fixed design choice, leaving its biologically and evolutionarily grounded dynamics underexplored.

In this work, we propose an evolutionary approach to developmental reward schedule discovery in RL that directly addresses this gap. Specifically, we seek to discover the functionally optimal developmental reward schedule for an RL agent, given a certain task, and we let evolution discover it --- as it did with biological organisms.

In our method, we define a composite reward function, inspired by motivational-cognitive development, comprising three biologically grounded components, each corresponding to a documented stage of maturation: agency \cite{heckhausen2010,white1959}, novelty \cite{kidd2012, kidd2015}, and reactivity \cite{galvan2010, cohen2010, casey2008}. The time-varying weights on these components are parameterized and optimized by an evolutionary outer loop, while agents are trained in an inner loop using Proximal Policy Optimization (PPO) \cite{schulman2017} on MiniGrid \cite{chevalierboisvert2023} environments. Namely, we evaluate four evolutionary algorithms: Covariance Matrix Adaptation Evolution Strategy (CMA-ES) \cite{hansen2023}, Exponential Natural Evolution Strategy (xNES) \cite{glasmachers2010}, Differential Evolution (DE) \cite{storn1997}, and Linear Population Size Reduction Success-History Based Adaptive Differential Evolution (L-SHADE) \cite{tanabe2014}, against an extrinsically motivated baseline (which serves as our main comparison point) and three additional hand-designed baselines. Critically, the evolutionary optimization receives no prior bias toward any particular developmental ordering.


This research offers a distinct perspective on RL reward schedule design, conveying that biologically grounded motivational architectures, when discovered through evolutionary optimization rather than prescribed by hand, can offer a principled and effective foundation for reward function design in deep reinforcement learning.

\section{Related Work}
Our work sits at the intersection of three research threads: computational accounts of developmental learning, intrinsic motivation in RL, and evolutionary methods for RL design --- each offering an important contribution towards the problem we address. Thus, we explore the related work across each of these three areas, and discuss how the identified gaps in the current literature are addressed by our proposed approach.

\subsection{Developmental Learning in Reinforcement Learning}
The RL community has begun translating developmental psychology and neuroscience knowledge into computational designs. An example of this can be visualized with Nussenbaum and Hartley's work \cite{nussenbaum2019}, which demonstrates that, when RL models are fit to human behavioral data, the best-fitting parameters differ systematically across age groups. In their findings, Nussenbaum and Hartley discuss that children's behavior is best captured by RL models with higher exploration temperatures and different learning rates; this differs from the lower exploration temperatures and more efficient outcome weighting that best captures adult behavior. Such findings suggest that staging the learning process of an RL agent may carry functional significance, an idea which Arditi et al. explore in their work on perceptual development emulation \cite{arditi2025}. By progressively expanding what the agent can observe during training, they show that a staged observation curriculum accelerates learning and improves final policy quality.

In addition to the body of work discussed, curriculum learning, though conceptually distinct, similarly recognizes that the sequence of experiences matters. By ordering tasks or training samples from simple to complex, Narvekar et al.\ \cite{narvekar2020} show that structuring what an agent trains on substantially improves learning efficiency and agent performance.

However, none of these works address what the agent is motivated by, and how that motivation should evolve across training --- which our method intends to uncover.

\subsection{Intrinsic Motivation in Reinforcement Learning}
If developmental work tells us that the timing of learning matters, the intrinsic motivation literature offers a natural set of building blocks, suggesting what to vary over time. For instance, researchers have developed intrinsic reward signals --- self-generated bonuses that encourage exploration independent of task success --- which take on several distinct forms: count-based visitation bonuses that reward visitation of less-frequent states \cite{tang2017}, novelty-based exploration bonuses measured through state-action pair discrepancy \cite{yang2024} or state discrepancy in learned latent spaces \cite{wang2024}, and self-attention curiosity that generates temporary bonuses, which fade as the agent masters a state \cite{hu2021}. 

Although each of these works captures a distinct facet of intrinsic motivation, none fully address how the weighting between intrinsic and extrinsic components should evolve across training --- a gap our work attempts to address through evolutionary optimization.

\subsection{Evolutionary Methods in Reinforcement Learning}
If evolutionary optimization is to serve as a motivational schedule discovery mechanism, a natural question is: Can it be successful in doing so? Prior work suggests it can be.

Across a range of deep learning (DL) applications and tasks, including deep RL, evolutionary algorithms have demonstrated their ability to find compelling solutions. One such example comes from Okada's work, which compares multiple classes of evolutionary methods and demonstrates that DE outperforms Genetic Algorithms and Evolution Strategies in neuroevolution (as it pertains to RL tasks), attributing DE's advantage to its adaptive balance between exploration and exploitation \cite{okada2023}. Moreover, Segal and Sipper demonstrate that dynamically combining genetic algorithms with novelty search outperforms either of these two strategies when taken alone, in deceptive reward environments \cite{segal2022}. The Proximal Evolutionary Strategy \cite{peng2024} additionally shows that combining CMA-ES with surrogate models and gradient-based local search reduces sample complexity, while remaining competitive with gradient-based deep RL methods.

Beyond policy optimization, Oh et al.\ \cite{oh2025} demonstrate that evolutionary meta-learning can surpass hand-designed RL algorithms, including PPO, on held-out benchmarks. Their research discovers the learning rule itself, while our proposed work applies the same principle, but to a different target; we seek to discover not how the agent learns, but what it is motivated by, and across what developmental arc.

To our knowledge, no prior work has applied evolutionary optimization to the discovery of developmental reward schedules, nor has any computational study directly compared evolved motivational trajectories to the sequence that biological development itself favors.

\section{Method}
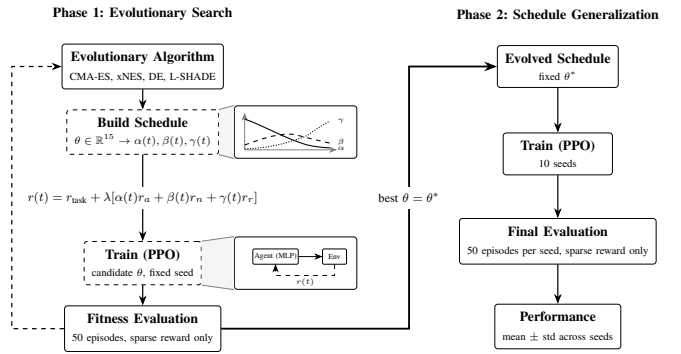
\begin{figure}[H]
\centering
\resizebox{\columnwidth}{!}{%
\begin{tikzpicture}[
  >=Stealth,
  every node/.style={font=\small},
  box/.style={draw, rounded corners=2pt, fill=white, minimum width=2.5cm, minimum height=0.8cm, align=center, inner sep=5pt},
  arr/.style={->, thick},
]
 
\node[font=\small\bfseries] at (-3, 1.2) {Phase 1: Evolutionary Search};
 
\node[box] (ea) at (-3, 0) {
  \textbf{Evolutionary Algorithm}\\[2pt]
  {\scriptsize CMA-ES, xNES, DE, L-SHADE}
};
 
\node[box, dashed, minimum width=2cm, minimum height=0.6cm] (dec) at (-3, -1.5) {
  \textbf{Build Schedule}\\[1pt]
  {\scriptsize $\theta \in \mathbb{R}^{15} \to \alpha(t), \beta(t), \gamma(t)$}
};
 
\node[box, dashed, minimum width=2cm, minimum height=0.6cm] (train) at (-3, -4.5) {
  \textbf{Train (PPO)}\\[1pt]
  {\scriptsize candidate $\theta$, fixed seed}
};
 
\node[box] (feval) at (-3, -6) {
  \textbf{Fitness Evaluation}\\[2pt]
  {\scriptsize 50 episodes, sparse reward only}
};
 
\draw[arr] (ea) -- (dec);
\draw[arr] (dec) -- (train);
\node[font=\footnotesize, fill=white, inner sep=3pt] at (-3, -3) {$r(t) = r_\text{task} + \lambda[\alpha (t) r_a + \beta (t) r_n + \gamma (t) r_r]$};
\draw[arr] (train) -- (feval);
\draw[arr, dashed] (feval.west) -- ++(-1.2, 0) |- (ea.west);
 
\node[font=\small\bfseries] at (6.5, 1.2) {Phase 2: Schedule Generalization};
 
\node[box] (best) at (6.5, 0) {
  \textbf{Evolved Schedule}\\[1pt]
  {\scriptsize fixed $\theta^*$}
};
 
\node[box] (train2) at (6.5, -2) {
  \textbf{Train (PPO)}\\[1pt]
  {\scriptsize 10 seeds}
};
 
\node[box] (eval2) at (6.5, -4) {
  \textbf{Final Evaluation}\\[2pt]
  {\scriptsize 50 episodes per seed, sparse reward only}
};
 
\node[box, minimum height=0.8cm] (report) at (6.5, -5.95) {
  \textbf{Performance}\\[2pt]
  {\scriptsize mean $\pm$ std across seeds}
};
 
\draw[arr] (best) -- (train2);
\draw[arr] (train2) -- (eval2);
\draw[arr] (eval2) -- (report);
 
\draw[arr, very thick] (feval.east) -- ++(4.3, 0) coordinate (corner) |- (best.west);
\node[font=\footnotesize, fill=white, inner sep=4pt] at ($(corner)!0.5!(corner |- best.west)$) {best $\theta = \theta^*$};
 
\node[draw, rounded corners=2pt, minimum width=2.8cm, minimum height=1.2cm] (dz) at (0.5, -1.5) {};
\fill[black!4]
  (dec.north east) -- (dz.north west) -- (dz.south west) -- (dec.south east) -- cycle;
\draw[gray, thin]
  (dec.north east) -- (dz.north west)
  (dec.south east) -- (dz.south west);
 
\begin{scope}
  \clip (dz.south west) rectangle (dz.north east);
  \begin{scope}[shift={($(dz.south west)+(0.24,0.2)$)}, xscale=0.3, yscale=0.22]
    \draw[->, line width=0.3pt, gray] (0,0) -- (7.0,0);
    \draw[->, line width=0.3pt, gray] (0,0) -- (0,3.8);
    \draw[line width=0.7pt]
      (0,3.2) .. controls (1.5,2.5) and (3,1.2) .. (4.8,0.5) .. controls (5.5,0.3) .. (6.5,0.2);
    \draw[line width=0.7pt, dashed]
      (0,0.5) .. controls (1,1.0) and (2.5,1.8) .. (3.8,1.6) .. controls (5.2,1.1) .. (6.5,0.7);
    \draw[line width=0.7pt, densely dotted]
      (0,0.1) .. controls (1.5,0.15) and (3,0.5) .. (4.8,1.5) .. controls (5.6,2.3) .. (6.5,3.0);
    \node[font=\tiny, anchor=west] at (6.7, 0.2) {$\alpha$};
    \node[font=\tiny, anchor=west] at (6.7, 0.85) {$\beta$};
    \node[font=\tiny, anchor=west] at (6.7, 3.0) {$\gamma$};
  \end{scope}
\end{scope}
 
\node[draw, rounded corners=2pt, minimum width=2.8cm, minimum height=1.2cm] (tz) at (0.5, -4.5) {};
\fill[black!4]
  (train.north east) -- (tz.north west) -- (tz.south west) -- (train.south east) -- cycle;
\draw[gray, thin]
  (train.north east) -- (tz.north west)
  (train.south east) -- (tz.south west);
 
\begin{scope}
  \clip (tz.south west) rectangle (tz.north east);
  \node[draw, rounded corners=1pt, minimum width=0.8cm, minimum height=0.35cm,
        font=\tiny, inner sep=1pt] (za) at ($(tz.center)+(-0.46,0.15)$) {Agent (MLP)};
  \node[draw, rounded corners=1pt, minimum width=0.55cm, minimum height=0.35cm,
        font=\tiny, inner sep=1pt] (ze) at ($(tz.center)+(0.89,0.15)$) {Env};
  \draw[->, thin] (za) -- (ze);
  \draw[->, thin, dashed] (ze.south) -- ++(0,-0.25) -| (za.south);
  \node[font=\tiny] at ($(tz.center)+(0.22,-0.43)$) {$r(t)$};
\end{scope}
 
\end{tikzpicture}%
}
\caption{Evolutionary reward schedule discovery framework}
\label{fig:pipeline}
\end{figure}

As we have established, a major challenge in reward schedule design is determining how the relative weights of motivational components should evolve across training. To address this challenge we propose a developmental-based, optimal reward schedule discovery framework comprising two distinct phases. The framework and its phases, as illustrated in Figure~\ref{fig:pipeline}, are described in the subsections that follow.

\subsection{Composite Reward Function}
We define a composite reward function comprising four components, three of which are grounded in distinct stages of biological motivational development, plus a fourth accounting for the sparse task reward:
\begin{equation}
r(t) = r_{\text{task}} + \lambda \bigl[\alpha(t) \cdot r_{\text{agency}} + \beta(t) \cdot r_{\text{novelty}} + \gamma(t) \cdot r_{\text{reactivity}}\bigr].
\end{equation}
Here, $r_{\text{task}}$ is the sparse environment reward computed as
\begin{equation}
  r_{\text{task}} =   
  \begin{cases}
  1 - 0.9 \cdot \dfrac{\text{steps}}{\text{steps}_{\max}} & \text{on success} \\[4pt]
  0 & \text{otherwise}
  \end{cases},
  \end{equation}
where steps is the number of steps taken by the agent within the episode and steps$_{\max}$ is the maximum number of steps allowed per episode, $\lambda = 0.003$ is a fixed intrinsic scale factor applied at every step, and $\alpha(t)$, $\beta(t)$, $\gamma(t)$ are time-varying, non-negative weights corresponding to agency, novelty, and reactivity, respectively. These weights govern the relative contribution of each motivational component over the course of training. Furthermore, the task reward is always present at full strength; evolution shapes only the relative contribution of the three developmental signals. 

\subsubsection*{Agency ($r_{\text{agency}}$)}Inspired by effectance motivation --- the earliest documented motivational drive \cite{heckhausen2010,white1959} --- this component rewards the agent whenever its action produces an observable state change:
\begin{equation}
r_{\text{agency}} = \begin{cases} 1 & \text{if } s_{t+1} \neq s_t \\ 0 & \text{otherwise} \end{cases},
\end{equation}
\noindent where $s_t$ is the agent's state at timestep $t$ and $s_{t+1}$ is the state following the agent's action. Specifically, the state $s$ is the tuple: \texttt{(agent position, facing direction, carrying status)}, and the reward fires whenever any of these three factors changes between consecutive steps.

This binary signal captures the fundamental satisfaction of causing effects in the environment, independent of what those effects may be. From the RL agent's perspective, it asks and answers the question: Can I make something happen? Which maps to the sensorimotor question an early-stage infant faces: Can my actions produce consequences in the world?

Moreover, the signal is non-decaying (it remains available throughout training) --- what changes is its weight in the composite reward function.

\subsubsection*{Novelty ($r_{\text{novelty}}$)} Inspired by mid-to-late-stage infants' documented attraction to novel stimuli, which diminishes as those stimuli become well-known \cite{kidd2012, kidd2015}, this component provides a count-based visitation bonus that decays with familiarity:
\begin{equation}
r_{\text{novelty}} = \frac{1}{\sqrt{N(s_t)}},
\end{equation}
\noindent where $N(s_t)$ is the number of times state $s_t$ has been visited. For novelty, the state $s$ additionally includes the status of each door (open or closed).

From the RL agent's perspective, it asks and answers the question: Have I been here before? --- which maps to the curiosity-driven question a growing infant faces: Is there more to discover beyond what I already know?

This signal is self-decaying by construction: as the agent revisits states, the bonus diminishes, naturally redirecting exploration toward less familiar regions of the state space.

\subsubsection*{Reactivity ($r_{\text{reactivity}}$)}

Motivated by a period of heightened reward sensitivity, namely the adolescent peak in striatal responsiveness preceding the maturation of the prefrontal cortex \cite{luciana2012,galvan2010,cohen2010,casey2008}, this component provides a dense, proximity-based signal that draws the agent toward the goal:
\begin{equation}
r_{\text{reactivity}} = 1 - \frac{d(s_t, g)}{d_{\max}},
\end{equation}
\noindent where $d(s_t, g)$ is the Manhattan distance from the agent to the goal and $d_{\max}$ is the maximum possible distance between any two cells in the grid.

From the RL agent's perspective, it asks and answers the question: How much continuous reward signal am I
receiving? Which maps to the appetitive question an adolescent faces: How quickly can I obtain the reward I desire?

Counter to agency and novelty, this signal is goal-directed by design. However, unlike the sparse task reward, which is received only upon goal completion, the reactivity signal provides continuous feedback at every timestep, drawing the agent closer to the goal but shaping the reward landscape throughout the episode rather than only at its conclusion. This design attempts to capture the impulsive pull an adolescent feels when pursuing a sought-after reward, and their strong preference for immediate over delayed gratification \cite{galvan2010,casey2008}.

\subsection{Weight Schedule Parameterization}
With the three motivational components outlined, we now address how their relative contributions should shift over the course of training. Rather than prescribing a fixed schedule, we parameterize each of the previously defined weight functions $\alpha(t)$, $\beta(t)$, $\gamma(t)$ as a piecewise linear function over $K$ evenly spaced control points. Here, $K = 5$, and each control point is defined by one unconstrained scalar parameter passed through a softplus activation function to enforce non-negativity. This parameterization yields a vector $\theta \in \mathbb{R}^{15}$, which serves as the search target for the evolutionary algorithm. Critically, the weights are not normalized during training --- the evolutionary algorithm is free to discover both the relative proportions and absolute magnitudes of each signal. Normalization summing to one is applied after training for visualization and reporting purposes.

\subsection{Phase 1: Evolutionary Search}
Phase one is responsible for optimal reward weight schedule discovery. Specifically, the question we are trying to solve with phase one is: Which schedule $\theta^* \in \mathbb{R}^{15}$ maximizes task performance, when used to train an agent from scratch?

To answer this question, we evaluate four evolutionary algorithms --- CMA-ES \cite{hansen2023}, xNES \cite{glasmachers2010}, DE \cite{storn1997}, and L-SHADE \cite{tanabe2014} --- each of which evolves a population of candidate schedule vectors across successive generations. For each candidate $\theta$, a PPO agent \cite{schulman2017} is trained from scratch, under the previously defined composite reward function $r(t)$, using a fixed training seed of 42 and an MLP policy with two hidden layers of 128 units. Upon completion, the agent is evaluated under the sparse task reward alone over 50 episodes, using a fixed evaluation seed of 123. The resulting mean episodic return is returned to the evolutionary algorithm as the fitness signal --- which the algorithm uses to update the population and propose new candidate schedules in the next generation. The loop repeats across 15 generations, after which the best-performing schedule $\theta^*$ is retained.

This design was formulated as so to ensure that evolution optimizes for task performance rather than for developmental signal maximization. 

\subsection{Phase 2: Schedule Generalization}
Phase two determines whether what was found in phase one is genuinely useful and generalizable, or merely a lucky artifact of a particular, random initialization.

In this phase, the best schedule $\theta^*$ resulting from the evolutionary search is held constant, and is evaluated by repeating PPO training independently across 10 seeds (seeds 42--51). Importantly, phase two follows the same per-run training and final evaluation protocol as phase one. Performance is reported as the mean $\pm$ the standard deviation of the final episodic return across seeds, as evaluated under the sparse task reward only.

\section{Experiments}
Our experimental setup and results compare evolved reward schedules against an extrinsically motivated baseline (serving as our principal comparison
point) and three additional hand-designed baselines, across two structured, goal-directed tasks, evaluating two criteria in particular: (1) whether evolution discovers more effective schedules than its non-evolved counterparts, and (2) whether these schedules mirror the motivational arc that biological development follows.

\subsection{Setup}
\subsubsection*{Environments}
We evaluate our framework on two MiniGrid \cite{chevalierboisvert2023} environments of varying complexity, both featuring sparse task rewards and procedurally randomized layouts across episodes:
  \begin{itemize}
       \item MiniGrid-DoorKey-6x6-v0: In a $6\times6$ fully observable grid, the agent must navigate the map, pick up a key, unlock the locked door, and reach the target square.
      \item MiniGrid-KeyCorridorS3R1-v0: In a fully observable corridor layout (three rooms joined by two doors), the agent must navigate the map, open an unlocked door, pick up a key, unlock the locked door, and pick up the target object (a ball).
  \end{itemize}

\subsubsection*{Conditions}
We compare four evolved conditions, two of which are Evolution Strategies (CMA-ES and xNES) and the remaining two which are Differential Evolution methods (DE and L-SHADE), against four baselines, for a total of eight distinct conditions:
  \begin{enumerate}
  \item CMA-ES: Weights optimized by CMA-ES
  \item xNES: Weights optimized by xNES
  \item DE: Weights optimized by DE
  \item L-SHADE: Weights optimized by L-SHADE
  \item Extrinsic Only: Uses the task reward alone with no added developmental signal
  \item Fixed Equal: All three developmental weights are held equal throughout training
  \item Developmental: Weights follow the biologically established developmental sequence: from agency, to novelty, to reactivity
  \item Reversed: The developmental schedule is applied in reverse order: from reactivity, to novelty, to agency
  \end{enumerate}

\subsubsection*{Hyperparameters}
The shared and algorithm-specific hyperparameters, used across all experiments, are summarized in Tables~\ref{tab:hyperparams} and~\ref{tab:algo_hyperparams}, respectively.

\begin{table}[h]
\caption{Shared Hyperparameters}
\label{tab:hyperparams}
\centering
\begin{tabular}{lll}
\toprule
\textbf{Component} & \textbf{Parameter} & \textbf{Value} \\
\midrule
\multirow{6}{*}{PPO}
  & Policy network      & MLP [128, 128] \\
  & Learning rate       & $3 \times 10^{-4}$ \\
  & Steps per update    & 512 \\
  & Batch size          & 64 \\
  & Epochs per update   & 4 \\
  & Discount $\gamma$   & 0.99 \\
\midrule
\multirow{5}{*}{Evolution}
  & Control points $K$  & 5 \\
  & Parameter dim       & 15 \\
  & Parameter bounds    & $[-10,\, 10]$ \\
  & Population size     & 12 \\
  & Max generations     & 15 \\
\midrule
\multirow{2}{*}{Seeds}
  & Phase 1 (search)         & 1 seed (seed 42) \\
  & Phase 2 (generalization) & 10 seeds (42--51) \\
\midrule
\multirow{3}{*}{Reward}
  & Intrinsic scale $\lambda$ & 0.003 \\
  & Eval episodes       & 50 \\
  & Eval seed           & 123 \\
\midrule
\multirow{2}{*}{Tasks}
  & DoorKey-6x6         & 2{,}000{,}000 timesteps \\
  & KeyCorridorS3R1     & 500{,}000 timesteps \\
\bottomrule
\end{tabular}
\end{table}

\begin{table}[h]
\caption{Algorithm-Specific Hyperparameters}
\label{tab:algo_hyperparams}
\centering
\begin{tabular*}{\columnwidth}{@{\extracolsep{\fill}}lll}
\toprule
\textbf{Algorithm} & \textbf{Parameter} & \textbf{Value} \\
\midrule
\multirow{1}{*}{CMA-ES}
  & Initial step size $\sigma_0$  & 2.0 \\
\midrule
\multirow{3}{*}{xNES}
  & Initial step size $\sigma_0$                        & 2.0 \\
  & Mean learning rate $\eta_\mu$                       & 1.0 \\
  & Step-size learning rate $\eta_{\sigma,B}$\textsuperscript{$\dagger$} & 0.295 \\
\midrule
\multirow{3}{*}{DE}
  & Strategy\textsuperscript{$\ddagger$} & best/1/bin \\
  & Mutation factor $F$ & 0.5 \\
  & Crossover rate $CR$ & 0.7 \\
\midrule
\multirow{3}{*}{L-SHADE}
  & History size $H$          & 6 \\
  & Min population $N_{\min}$ & 4 \\
  & Top fraction $p$          & 0.1 \\
\bottomrule
\end{tabular*}
\vspace{2pt}
\begin{minipage}{\columnwidth}
\footnotesize
$^{\dagger}$ Computed as $(3+\ln d)/(5\sqrt{d})$, where $d=15$.\\[1pt]
$^{\ddagger}$ best/1/bin denotes best-candidate base vector, one difference vector, and the binomial crossover, respectively.
\end{minipage}
\end{table}

\subsubsection*{Hardware}
All experiments were run on a single Apple M3 Max chip (14-core CPU, 36 GB of unified memory); evolutionary fitness evaluations were parallelized across CPU cores.

\subsection{Findings}
\subsubsection*{Search Convergence}
Across all runs, the evolutionary search settled well within its 15-generation budget. On DoorKey-6x6, the best-so-far fitness improved over the first several generations before plateauing by generation 7 at the latest, and on KeyCorridorS3R1, the fitness plateau occurred within the initial population (generation 0).

\subsubsection*{Performance}
On DoorKey-6x6, all four evolutionary algorithms outperform the extrinsic only baseline, which achieves a mean return of $0.668$, with L-SHADE achieving the highest mean return of $0.744$, followed by DE ($0.730$), then CMA-ES ($0.714$), and lastly xNES ($0.672$).

On KeyCorridorS3R1, the results differ considerably. For this task, CMA-ES achieves the highest mean return of $0.902$, with the extrinsic only baseline following closely behind, achieving a competitive mean return of $0.871$. In third place, xNES achieves a moderate mean return of $0.755$. DE and L-SHADE, by contrast, show high variance and overall poor performance, with both methods obtaining the same mean return of $0.266$. Notably, for seven of the ten evaluated seeds, these two algorithms converged to degenerate schedules (episodic returns of 0).

\begin{figure}[h]
  \centering
  \includegraphics[width=\columnwidth]{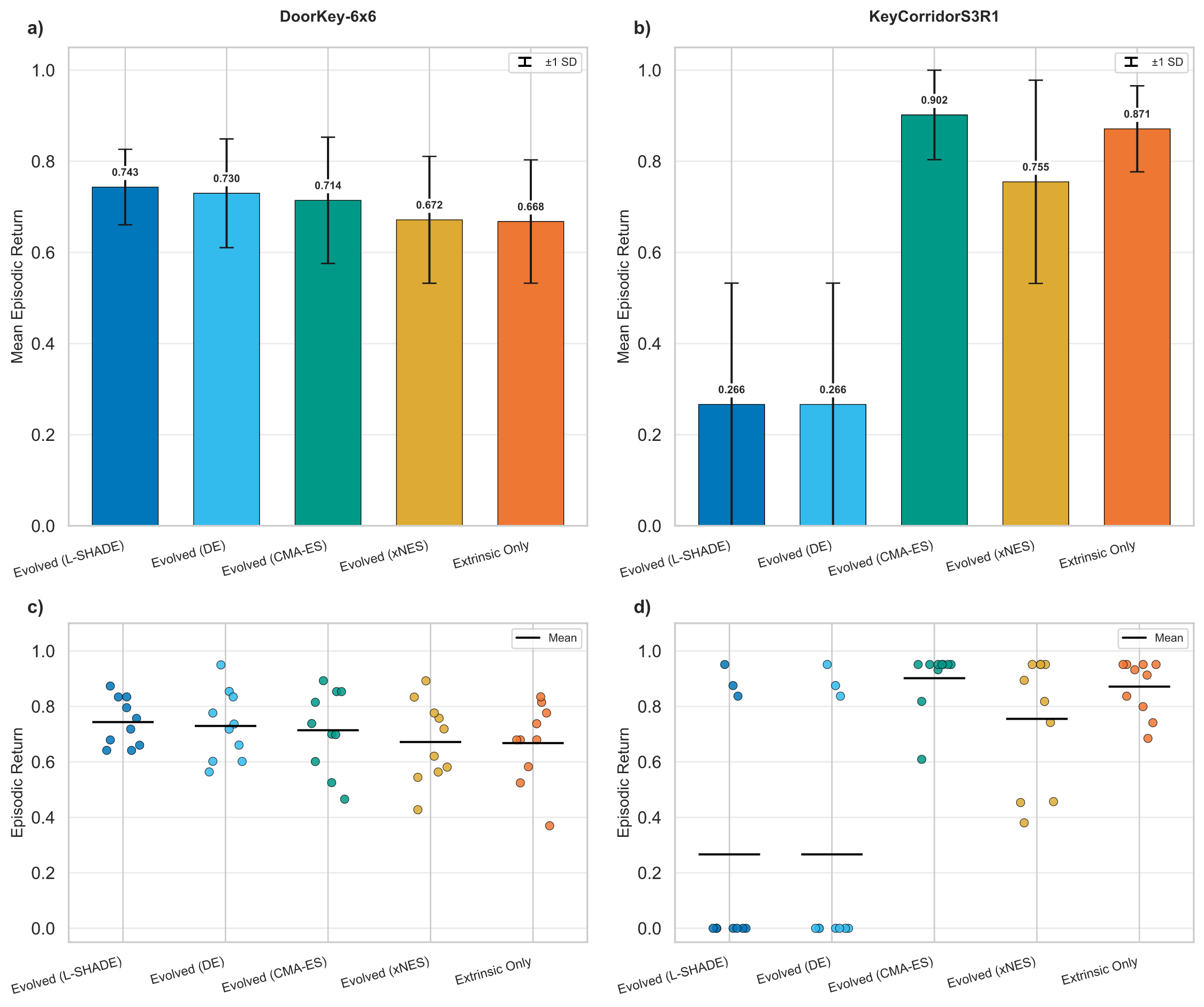}
  \caption{Performance comparison across all evolved conditions, plus the extrinsic only baseline, on both tasks. \textbf{(a)} Mean episodic return $\pm$1\,SD, across 10 seeds, for DoorKey-6x6. \textbf{(b)} Mean episodic return $\pm$1\,SD, across 10 seeds, for KeyCorridorS3R1. \textbf{(c)} Per-seed returns with mean lines for DoorKey-6x6. \textbf{(d)} Per-seed returns with mean lines for KeyCorridorS3R1.}
  \label{fig:fig1}
\end{figure}

All of these performance results are summarized in Figure~\ref{fig:fig1} and Table~\ref{tab:results}.

\begin{table}[H]
\caption{Mean Episodic Return $\pm$ Std --- Evolved Conditions}
\label{tab:results}
\centering
\begin{tabular}{lcc}
\toprule
\textbf{Condition} & \textbf{DoorKey-6x6} & \textbf{KeyCorridorS3R1} \\
\midrule
L-SHADE   & $\mathbf{0.744 \pm 0.083}$ & $0.266 \pm 0.408$ \\
DE       & $\underline{0.730 \pm 0.119}$ & $0.266 \pm 0.408$ \\
CMA-ES    & $0.714 \pm 0.139$ & $\mathbf{0.902 \pm 0.105}$ \\
xNES      & $0.672 \pm 0.139$ & $0.755 \pm 0.223$ \\
\midrule
Extrinsic only      & $0.668 \pm 0.135$ & $\underline{0.871 \pm 0.094}$ \\
\bottomrule
\end{tabular}
\end{table}

\subsubsection*{Evolved Weight Schedules}
\begin{figure}[h]
  \centering
  \includegraphics[width=\columnwidth]{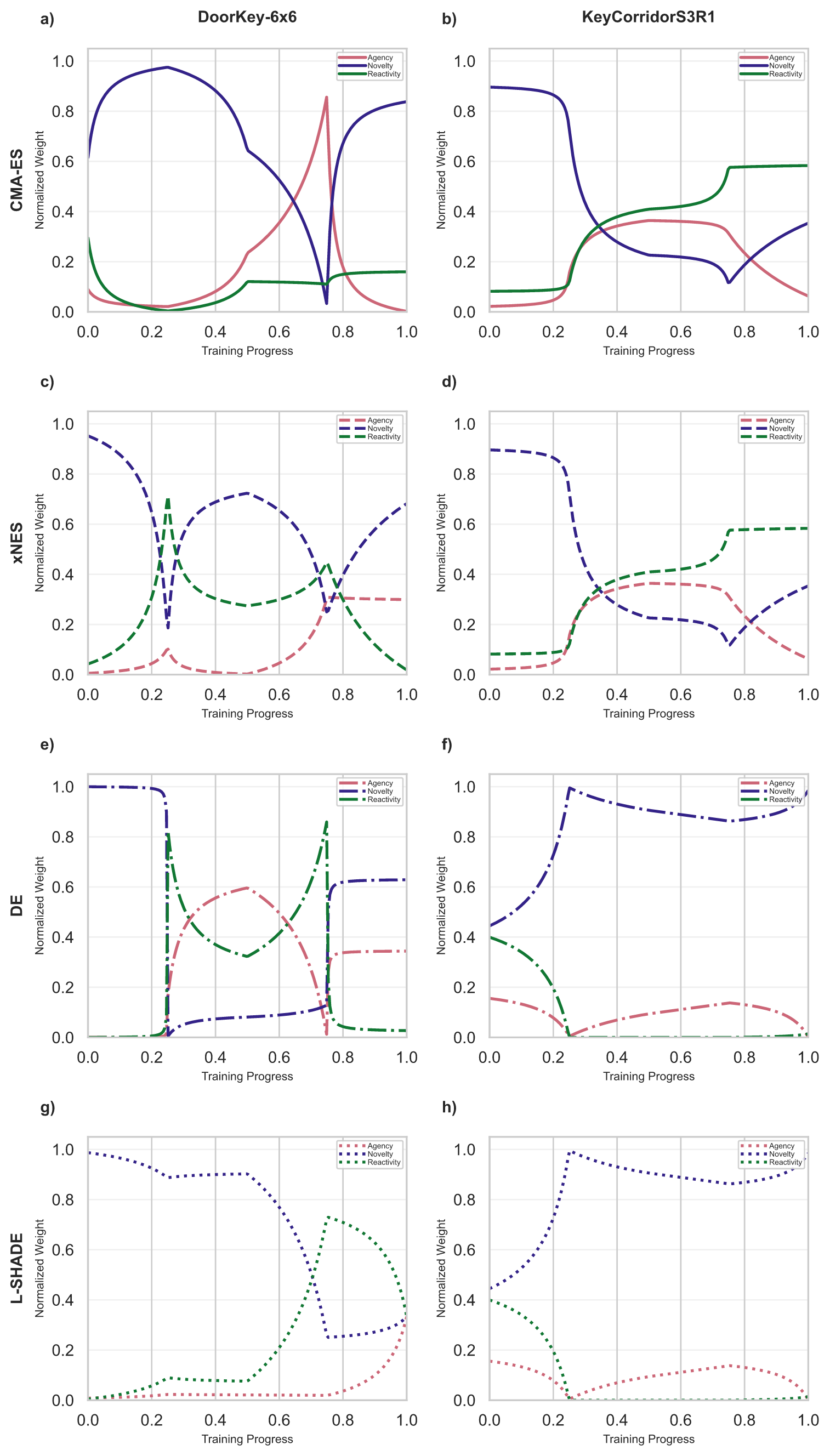}
  \caption{Evolved weight schedules for the four evolutionary algorithms, across both tasks.
  \textbf{Rows:} CMA-ES (a--b), xNES (c--d), DE (e--f), L-SHADE (g--h). \textbf{Columns:} DoorKey-6x6 (left), KeyCorridorS3R1 (right). Plotted weights are normalized to sum to one at each timestep, rendering the underlying piecewise-linear weights as smooth curves.}
  \label{fig:fig3}
\end{figure}

The evolved weight schedules varied substantially across tasks and algorithms, as shown in Figure~\ref{fig:fig3}. Notably, for both tasks, all algorithms begin their training with novelty as the dominant signal, reflecting an early emphasis on exploration. Past these initial stages, distinct behavior emerges.

On DoorKey-6x6, CMA-ES, xNES, and DE follow complex trajectories that all end with novelty as the highest-weighted component. L-SHADE is the
exception, transitioning from novelty-dominant to reactivity-dominant weighting in the final third of training. This divergence is particularly interesting, given that L-SHADE is the best performing condition for this task.

On KeyCorridorS3R1, CMA-ES and xNES converge to identical schedules: novelty dominates early before giving way to reactivity. DE and L-SHADE likewise converge to equivalent schedules, where novelty rises rapidly early and maintains this dominant behavior for the remainder of training.

\subsubsection*{Learning Dynamics}
\begin{figure}[H]
  \centering
  \includegraphics[width=\columnwidth]{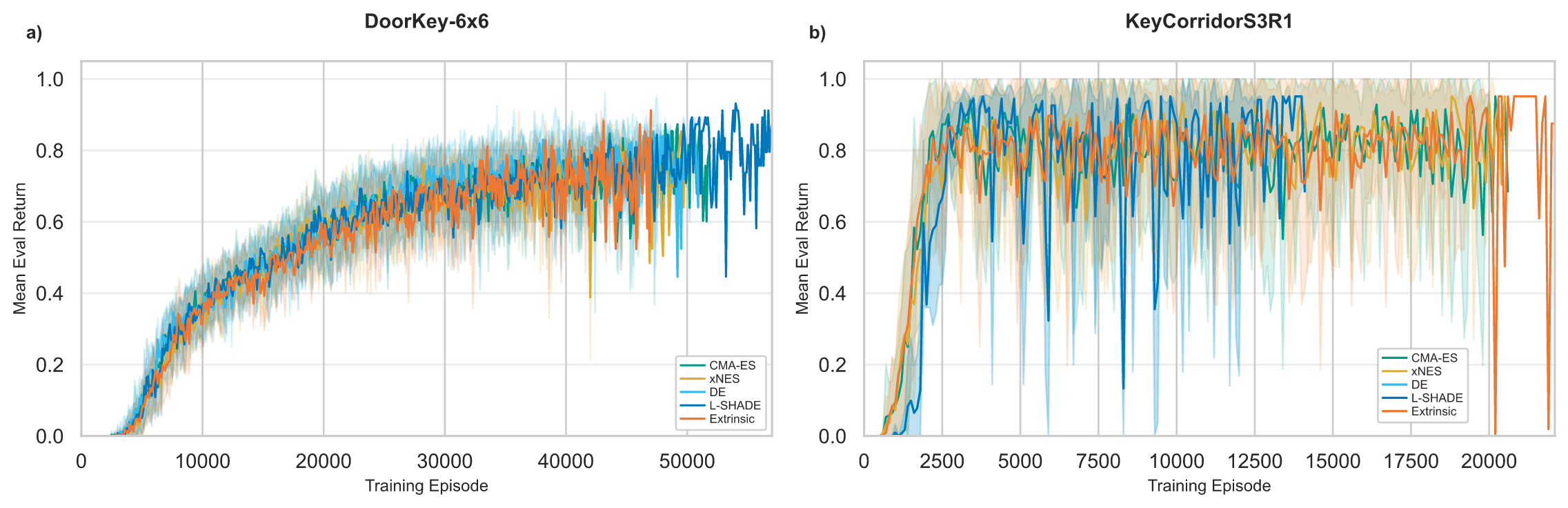}
  \caption{Learning curves for all evolved conditions plus the extrinsic only baseline, across both tasks. Mean evaluation return $\pm$1\,SD (shaded) over 10 seeds, plotted against training episode. \textbf{(a)} DoorKey-6x6. \textbf{(b)} KeyCorridorS3R1. The x-axis spans the longest training run across all methods for each task; methods with shorter training horizons terminate earlier.}
  \label{fig:fig5}
\end{figure}

As part of our experimental framework, we tracked learning dynamics across training, which allowed us to better understand why certain algorithms fail to generalize.
Figure~\ref{fig:fig5} shows these learning dynamics.

On DoorKey-6x6, CMA-ES, xNES, DE, and L-SHADE all exhibit stable, increasing learning curves that converge within the training budget. As for KeyCorridorS3R1, CMA-ES and xNES converge reliably, while DE and L-SHADE display unusually high inter-seed variance, with their mean curves being pulled downward by the seven seeds, per algorithm, that converged to degenerate solutions.

\subsubsection*{Agent Behavior}
\begin{figure}[h]
  \centering
  \includegraphics[width=\columnwidth]{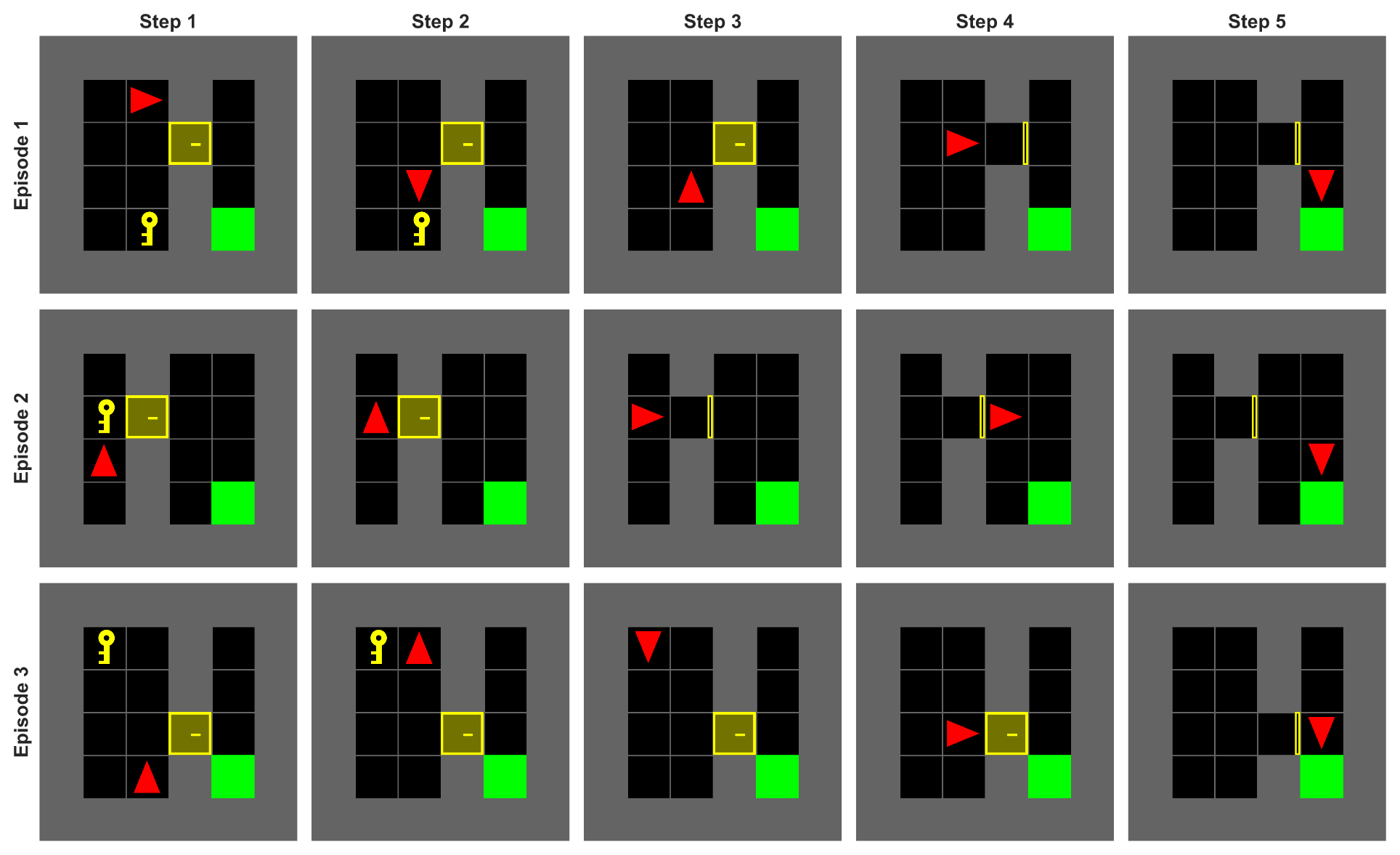}
  \caption{Best evolved agent (L-SHADE, seed 46) solving DoorKey-6x6. Three representative episodes \textbf{(rows)} across five evenly sampled timesteps \textbf{(columns)}.}
  \label{fig:fig6_doorkey}
\end{figure}

\begin{figure}[h]
  \centering
  \includegraphics[width=\columnwidth]{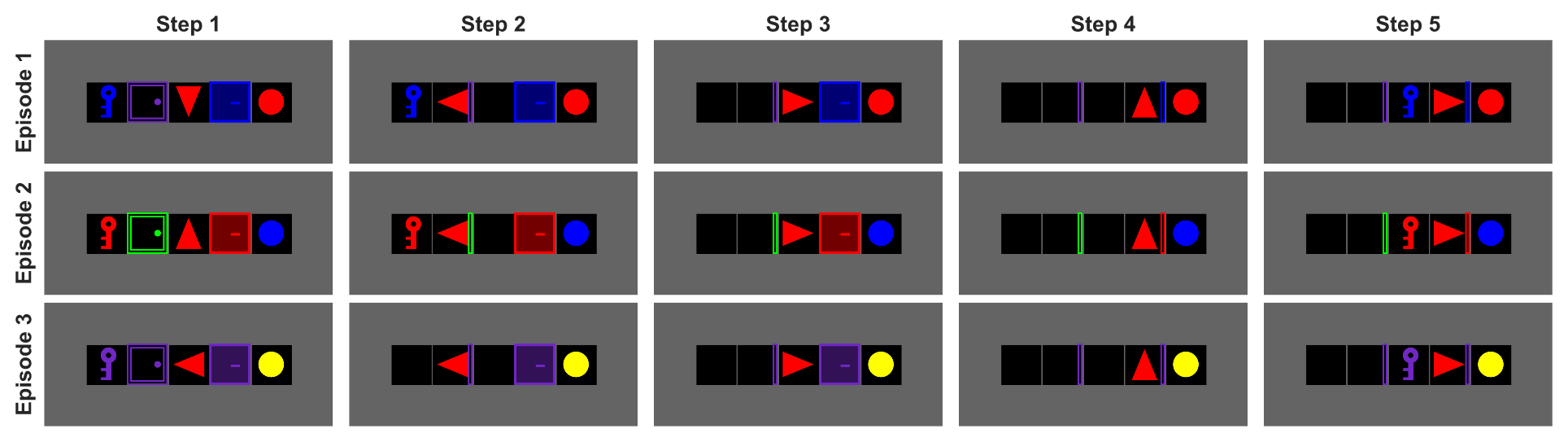}
  \caption{Best evolved agent (CMA-ES, seed 42) solving KeyCorridorS3R1. Three representative episodes \textbf{(rows)} across five evenly sampled timesteps \textbf{(columns)}.}
  \label{fig:fig6_keycorridor}
\end{figure}

The best evolved agent, for each respective task, demonstrates that the discovered schedules produce coherent, goal-directed behavior, as visualized in Figure~\ref{fig:fig6_doorkey} (DoorKey-6x6) and Figure~\ref{fig:fig6_keycorridor} (KeyCorridorS3R1).

\subsection{Ablations}
\begin{figure}[h]
  \centering
  \includegraphics[width=\columnwidth]{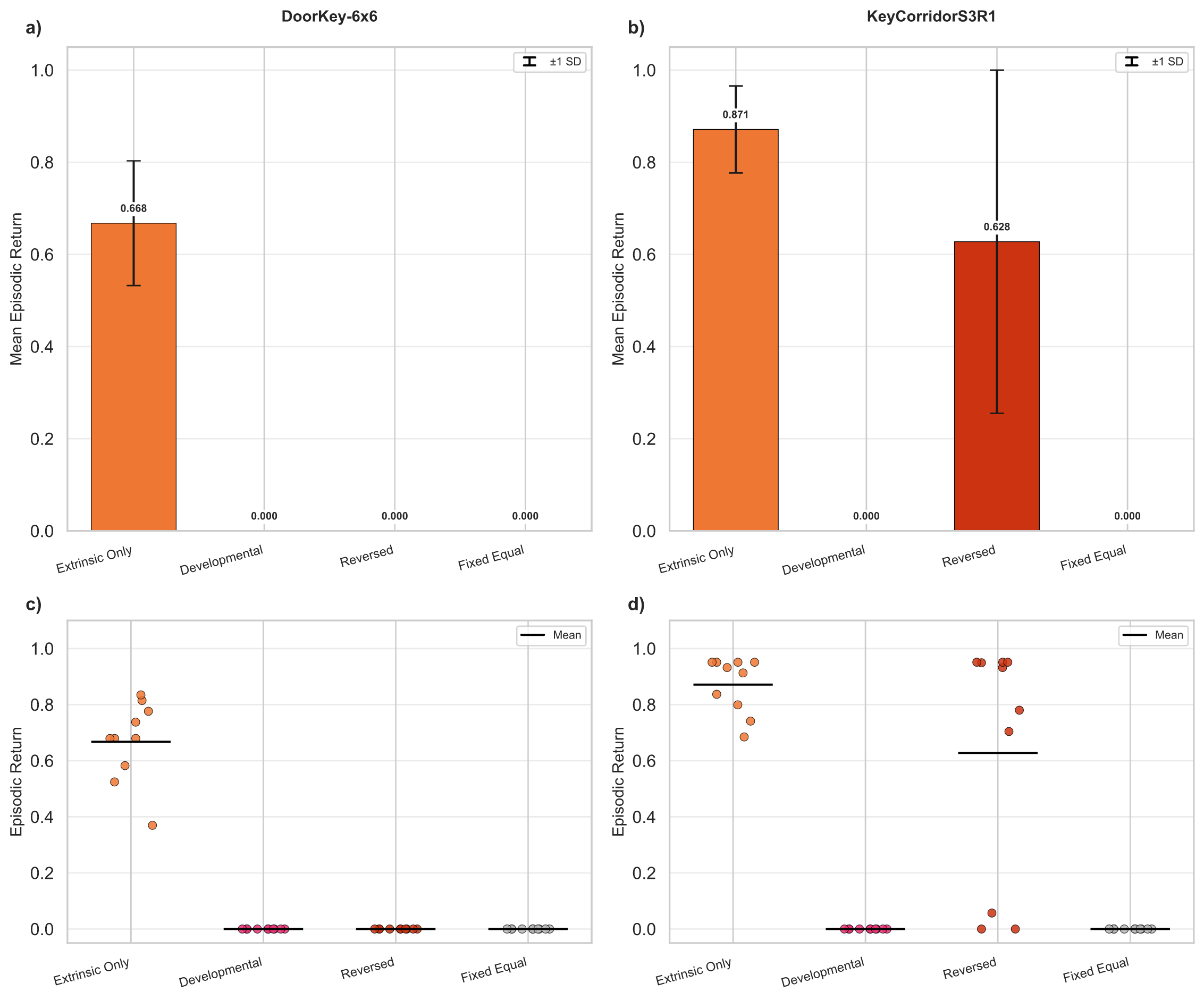}
  \caption{Performance of all baseline conditions (including ablations). \textbf{(a)} Mean episodic return $\pm$1\,SD, across 10 seeds, for DoorKey-6x6. \textbf{(b)} Mean episodic return $\pm$1\,SD, across 10 seeds, for KeyCorridorS3R1. \textbf{(c)} Per-seed returns with mean lines for DoorKey-6x6. \textbf{(d)} Per-seed returns with mean lines for KeyCorridorS3R1.}
  \label{fig:fig1_ablations}
\end{figure}

The ablation conditions tested are designed to isolate the factors that make the evolved schedules effective. Three competing hypotheses are tested:
\begin{itemize}
    \item Fixed Equal
    \begin{itemize}
      \item Setup: Asks whether the advantage comes simply from having multiple simultaneous developmental signals, rather than from their temporal ordering.
      \item Findings: Its zero return on both tasks rules this out --- schedule dynamics matter, not just signal diversity.
    \end{itemize}
    \item Developmental
    \begin{itemize}
      \item Setup: Asks whether the biologically motivated ordering, applied as a fixed linear schedule, is sufficient on its own (and computationally optimal).
      \item Findings: Its failure on both tasks shows it is not sufficient nor computationally optimal --- per-task adaptation is necessary.
    \end{itemize}
    \item Reversed
    \begin{itemize}
      \item Setup: Asks whether the direction of the developmental sequence matters, or whether any ordered transition between motivational components is beneficial.
      \item Findings: It achieves zero on DoorKey-6x6 and only partial generalization on KeyCorridorS3R1 ($0.628 \pm 0.407$), showing that the ordering of motivational components matters during learning.
    \end{itemize}
  \end{itemize}
These results are summarized in Table~\ref{tab:ablations} and illustrated, across all conditions and seeds, in Figure~\ref{fig:fig1_ablations}.

\begin{table}[h]
\caption{Mean Episodic Return $\pm$ Std --- Baseline Conditions}
\label{tab:ablations}
\centering
\begin{tabular}{lcc}
\toprule
\textbf{Condition} & \textbf{DoorKey-6x6} & \textbf{KeyCorridorS3R1} \\
\midrule
Developmental   & $0.000 \pm 0.000$ & $0.000 \pm 0.000$ \\
Reversed        & $0.000 \pm 0.000$ & $\underline{0.628 \pm 0.407}$ \\
Fixed Equal     & $0.000 \pm 0.000$ & $0.000 \pm 0.000$ \\
\midrule
Extrinsic Only  & $\mathbf{0.668 \pm 0.135}$ & $\mathbf{0.871 \pm 0.094}$ \\
\bottomrule
\end{tabular}
\end{table}

\section{Discussion}
Returning to the two core criteria outlined in the experiments, the results demonstrate that evolved schedules are effective, though on some tasks more than others, but do not necessarily mirror the biological motivational arc. 

Regarding the effectiveness of evolved schedules, we can confidently assert their outperformance compared to all non-evolved alternatives on DoorKey-6x6. On KeyCorridorS3R1, CMA-ES similarly outperforms all baselines, though the other algorithms fall short of the extrinsic only method, with DE and L-SHADE being additionally surpassed by the reversed baseline condition. 

We hypothesize that the observed performance difference stems from: (1) how task difficulty shapes the search, and (2) the type of evolutionary algorithm family being evaluated. On DoorKey-6x6 (the harder task), the extrinsic reward alone is insufficient, so performance depends more heavily on the evolved schedule; here, the fitness landscape is informative, fitness improves over generations for all four methods, and the resulting schedules generalize across seeds (with the best performers being L-SHADE and DE). On KeyCorridorS3R1, however, the extrinsic reward alone already performs strongly across seeds. Thus, schedules attributing higher weights to developmental components risk converging to a niche or seed-specific solution which fails to generalize well --- as was the case with L-SHADE and DE. Conversely, CMA-ES and xNES attributed lower weights to the three developmental components, and as a result were able to generalize better across seeds, relative to the two other evolutionary methods. Interestingly, CMA-ES and xNES both fall within the Evolution Strategies family of evolutionary algorithms, while L-SHADE and DE fall within the Differential Evolution family. Our findings, thus, support the framing that the Evolution Strategies family is better suited for easier tasks, while the Differential Evolution family performs better on harder ones, and suggest that evolutionary optimization, as applied to developmental-based RL schedule discovery, can be an effective method depending on the task at-hand.

As it pertains to biological alignment, the discovered schedules appear to be distinct to the biological ideal observed throughout mammalian development. The biological developmental sequence: effectance motivation in the neonatal period \cite{heckhausen2010,white1959}, curiosity and novelty-seeking behavior in mid-to-late infancy \cite{kidd2012}, and heightened reward sensitivity in adolescence \cite{galvan2010,cohen2010} predicts an agency-to-novelty-to-reactivity ordering. Yet, all four evolved algorithms begin training with novelty, across both tasks, as the dominant signal --- and notably none recover or substantially align with the full biological sequence.

Although this finding suggests that, in terms of optimal developmental-based reward system discovery, biological evolution and computational evolution do not speak the same language, it is important to consider that this experimental setting represents a vastly simplified interpretation of both motivational-cognitive development and evolutionary biology, and thus the observed divergence may be specific to the proposed tasks and reward formulations.

\subsection{Limitations}
We note the following limitations, each of which, if addressed, could strengthen this research. Furthermore, these limitations provide future directions for this work.

First, the intrinsic scale $\lambda = 0.003$ was fixed rather than evolved; allowing the scale to vary may result in distinct, improved schedules. Second, we only evaluated our method on two grid-world tasks with discrete action spaces; continuous-control, partially observable, or other more complex environments may result in different evolved schedules and results. Third, the reward components are hand-designed proxies of biological drives; strict derivations from computational neuroscience models could yield more principled and potentially better results. Fourth, statistical significance is omitted, as the current sample size of $n=10$ seeds may be insufficient to reliably detect it, given the effect sizes observed; a larger sample would provide more accurate hypothesis testing. Fifth, phase one identifies a single optimal schedule, which is then utilized to test the effect of generalization; optimal schedules may vary across seeds, with individual schedule discovery potentially yielding stronger per-seed evaluation performance compared to our tested approach. Sixth, removing each developmental signal individually would further isolate their unique contributions toward the evolved schedule's effectiveness; an additional ablation experiment could provide this insight. Finally, each fitness evaluation requires a full PPO training run; more sample-efficient meta-learning approaches \cite{oh2025} could make our framework more broadly applicable. 

\subsection{Future Work}
Several directions follow naturally from these findings and the outlined limitations. As a first approach, per-seed schedule discovery in phase one would allow us to examine whether optimal schedules vary across random initializations, and whether evolved schedules consistently outperform baselines when the seed used for training is dynamic rather than fixed. Moreover, evolving the intrinsic scale $\lambda$ jointly with the weight schedule offers an interesting yet simple way to potentially increase the performance of our agent, and discover different reward schedules. Lastly, evaluating the framework on a broader range of environments with distinct settings and varying degrees of complexity would allow us to test the extent to which our framework is task-dependent.

\section{Conclusion}
We presented a framework for discovering developmental reward schedules via evolutionary optimization, and demonstrated that evolved schedules can outperform fixed-weight and other hand-designed alternatives across two structured, goal-directed tasks. On DoorKey-6x6, all four tested algorithms favor novelty as the dominant early signal and outperform all baselines, with L-SHADE standing out as the best performer. On KeyCorridorS3R1, CMA-ES surpassed all other conditions, though the other three evolved methods fell short, underperforming the extrinsic only baseline. Here, DE and L-SHADE frequently converge to degenerate solutions. Critically, fixed developmental schedules --- despite similarly being biologically motivated --- actively interfere with task learning, underscoring the importance of per-task adaptation, which evolutionary optimization facilitates. Moreover, our findings reveal that, within the bounds of our tasks and framework, computational and biological evolution do not discover the same motivational sequence.

In summary, this work positions evolutionary optimization as a promising method for developmental reward schedule discovery, and contributes a novel perspective on how developmental principles can inform, and be challenged by, reward design in deep reinforcement learning.

\section*{ACKNOWLEDGMENT}
The author declares no funding sources for this research. AI tools were used to assist with the ideation and brainstorming, coding, and manuscript preparation processes. All AI-generated content was screened and verified by the author prior to inclusion in the manuscript.

\end{document}